\documentclass[runningheads]{llncs}
\usepackage[T1]{fontenc}
\usepackage{booktabs}    % for \toprule, \midrule, \bottomrule
\usepackage{multirow}    % for multirow cells
\usepackage{array}       % for better column alignment
\usepackage{algorithm}
\usepackage[table]{xcolor}

\definecolor{LightPurple}{gray}{0.93}
\usepackage{algorithmicx}
\usepackage{algpseudocode}
\usepackage{pifont}
\newcommand{\cmark}{\ding{51}}  % ✔
\newcommand{\xmark}{\ding{55}}  % ✗

\usepackage{multirow}
\usepackage[T1]{fontenc}
\usepackage{marvosym}

\usepackage{graphicx}
\usepackage{amsmath,amssymb,amsfonts}
\usepackage{booktabs}
\usepackage{bm}
\usepackage{hyperref}
\usepackage{cite}
\usepackage{amssymb} % 用于 \checkmark
\usepackage{pifont}  % 用于 \xmark (可选)

\usepackage{colortbl} % 用于背景色高亮
\definecolor{graybg}{gray}{0.95}
\usepackage[T1]{fontenc}
\usepackage{graphicx,verbatim}
\usepackage{graphicx,verbatim}
\begin{document}
\title{Hallucination Detection and Correction in Medical VLMs via Counter-Evidence Verification}
\begin{comment}  %% Removed for anonymized MICCAI submission
\author{First Author\inst{1}\orcidID{0000-1111-2222-3333} \and
Second Author\inst{2,3}\orcidID{1111-2222-3333-4444} \and
Third Author\inst{3}\orcidID{2222--3333-4444-5555}}
%
\authorrunning{F. Author et al.}
\institute{Princeton University, Princeton NJ 08544, USA \and
Springer Heidelberg, Tiergartenstr. 17, 69121 Heidelberg, Germany
\email{lncs@springer.com}\\
\url{http://www.springer.com/gp/computer-science/lncs} \and
ABC Institute, Rupert-Karls-University Heidelberg, Heidelberg, Germany\\
\email{\{abc,lncs\}@uni-heidelberg.de}}

\end{comment}

\author{Nan Zhou\inst{1}\index{Zhou, Nan}
\and
Ke Zou\inst{2}\index{Zou, Ke}
\and
Meng Liu\inst{1}\index{Liu, Meng}
\and
Linchao He\inst{1}\index{He, Linchao}
\and
Jiaqi Zhu\inst{4}\index{Zhu, Jiaqi}
\and
Yi Zhang\inst{3}\index{Zhang, Yi}
\and \\
Hu Chen\inst{1}$^{\textrm{\Letter}}$\index{Chen, Hu}
\and
Huazhu Fu\inst{5}$^{\textrm{\Letter}}$\index{Fu, Huazhu}
}

\authorrunning{Nan Zhou et al., submission to MICCAI 2026 review}
\institute{the College of Computer Science, Sichuan University. \and
% the Centre for Innovation \& Precision Eye Health, Department of Ophthalmology,
Yong Loo Lin School of Medicine, National University of Singapore.\and
the School of Cyber Science and Engineering and the Key Laboratory of Data Protection and Intelligent Management, Ministry of Education, Sichuan University,  \and
 National Key Laboratory of Autonomous Intelligent Unmanned Systems, Beijing Institute of Technology. \and
 the Institute of High Performance Computing (IHPC), Agency for Science, Technology and Research (A*STAR).  \\
\email{huchen@scu.edu.cn; hzfu@ieee.org} \\
$^{\textrm{\Letter}}$Corresponding authors.}
  
\maketitle              % typeset the header of the contribution

\begin{abstract}
% The reliability of vision-language models (VLMs) remains a major challenge in medical diagnosis, largely due to hallucinations that undermine clinical trust. 
Vision-Language models (VLMs) reliability in medical diagnosis is challenged by trust-undermining hallucinations.
Existing hallucination detection approaches mainly focus on identifying factual inconsistencies between generated text and reference data. While some studies analyze where models attend in images, they seldom verify whether such attention truly reflects the visual evidence supporting the generated text. To address this gap, we propose \textbf{{Co}unter-{E}vidence{ V}erification ({CoEV})}, a training-free plug-and-play framework that detects and corrects hallucinations through evidence-based factual consistency verification. CoEV performs bidirectional verification between textual assertions and visual evidence, testing whether each statement is supported by its corresponding evidence region, and assigns each statement into a four-quadrant diagnostic map capturing combinations of text factuality and visual grounding.
CoEV detects hallucinated content and serves as a post hoc refinement tool, correcting hallucinations without retraining. Extensive experiments on four medical datasets show that CoEV combats hallucinations in VLMs.For hallucination detection, CoEV consistently outperforms existing methods, improving average PR-AUC and ROC-AUC by 3.0\% and 3.9\% absolute points respectively, with notable gains of up to 18.5\% in specific VQA scenarios. For hallucination correction, it improves Micro-F1 by up to 12.5\%, reduces hallucination rates by over 11.9\% on medical report generation, and also boosts medical VQA accuracy. These results show that CoEV enables reliable detection and correction of hallucinations, providing clinicians with dependable, evidence-based cues for diagnosis. Code will be released upon acceptance.

% Authors must provide keywords and are not allowed to remove this Keyword section.

\end{abstract}
\section{Introduction}
Medical Vision-Language Models (VLMs)~\cite{zhang2023biomedclip,li2023llava,sellergren2025medgemma,thawkar2023xraygpt} have demonstrated transformative potential in clinical workflows, enabling automated interpretation across diverse tasks such as Medical Visual Question Answering (Med-VQA), Radiology Report Generation (MRG), and precise phrase grounding~\cite{zou2025uncertainty,bannur2024maira,liu2025gemex,zou2024medrg}. By bridging high-dimensional visual features with complex clinical vocabulary, these models facilitate rapid diagnostic assistance. However, the deployment of VLMs in high-stakes clinical environments remains constrained by persistent hallucinations, where models generate clinically inaccurate, contradictory, or fabricated descriptions that lack support from underlying visual evidence~\cite{liu2024survey}. In the medical domain, such ungrounded outputs are not merely linguistic errors; they represent significant safety risks, causing erroneous treatment recommendations and undermining clinician trust in AI-assisted decision-making~\cite{liu2024survey}.

% \begin{figure}[t]
%     \centering
%     \includegraphics[width=1\linewidth]{F-0_V1.pdf} % 替换为你的图文件
%     % \includegraphics[width=1\linewidth]{sec/Fig/F-2.png} % 替换为你的图文件

%     % \caption{(a)\textbf{ Overview of the CoEV framework.} The {CoEV diagnostic process} performs counter-evidence verification and four-quadrant mapping to identify visually grounded and hallucinated claims from VLM outputs. (b) \textbf{CoEV for Med-VQA:} diagnostic signals are used to verify answer validity and identify hallucinated predictions. (c) \textbf{CoEV for Medical Report Refinement:} detected hallucinated sentences are refined through CoEV-guided rewriting.}
%    \caption{
% (a) Traditional factual verification methods for VLMs fall into two categories (“What \& Where”). The ``What'' category checks textual factuality, and the ``Where'' category links text to image regions.
% (b)The Four-Quadrant Diagnostic Map. Defined by a textual axis (factuality) and a visual axis (grounding), the map comprises four quadrants: Consistent-Grounded, Consistent-Ungrounded, Inconsistent-Grounded, and Inconsistent-Ungrounded.
% % (c) The four-quadrant diagnostic space. In this framework, the { Consistent-Grounded} corresponds to textual claims that are fully supported by visual evidence in the image, and so on for all quadrants. %the \textbf{Consistent-Ungrounded} includes claims that appear plausible but lack sufficient support from the image,
% }
%     \label{fig:F-2}
% \end{figure}
Existing methods for hallucination detection primarily leverage uncertainty estimation, cross-model consistency checks~\cite{farquhar2024detecting,moslonka2025learned,liao2025vision,zhang2024radflag}, or traditional linguistic similarity metrics like BLEU~\cite{papineni2002bleu} and ROUGE~\cite{lin2004rouge}. While these approaches provide a coarse-grained measure of output quality, they are often insensitive to the factual grounding required for medical accuracy. Specifically, textual similarity metrics fail to capture nuanced hallucinations arising from flawed visual reasoning, where a report may be fluently written but medically incorrect~\cite{farquhar2024detecting,moslonka2025learned}. On the other hand, interpretability tools such as attention visualization or gradient-based saliency maps provide spatial intuition~\cite{raghavan2024attention,an2022attention}; however, they merely highlight statistical correlations and do not verify the rigorous factual alignment between specific textual assertions and localized image regions. Consequently, there remains a need for a diagnostic framework that transcends simple detection to provide traceable evidence, identifying not only the presence of hallucinations but also localizing the underlying visual regions responsible for them.

To address this, we propose Counter-Evidence Verification (CoEV), a plug-and-play and training-free framework for detecting and correcting hallucinations in medical VLMs. Unlike methods requiring external models or fine-tuning~\cite{xiao2025detecting,gunjal2024detecting}, CoEV performs diagnosis-informed interventions by selectively masking visual regions as counterfactual probes. This tests whether assertions depend on visual evidence or linguistic priors. By analyzing behavioral shifts, CoEV introduces a Four-Quadrant Diagnostic Map to systematically categorize claims into states of factual consistency. As a post-hoc mechanism, CoEV enables effective mitigation during inference without retraining. Evaluations across four datasets and multiple VLMs demonstrate that CoEV significantly improves diagnostic accuracy and consistency, providing an interpretable pathway for clinical AI.

\begin{figure}[ht]
    \centering
    \includegraphics[width=0.92\linewidth]{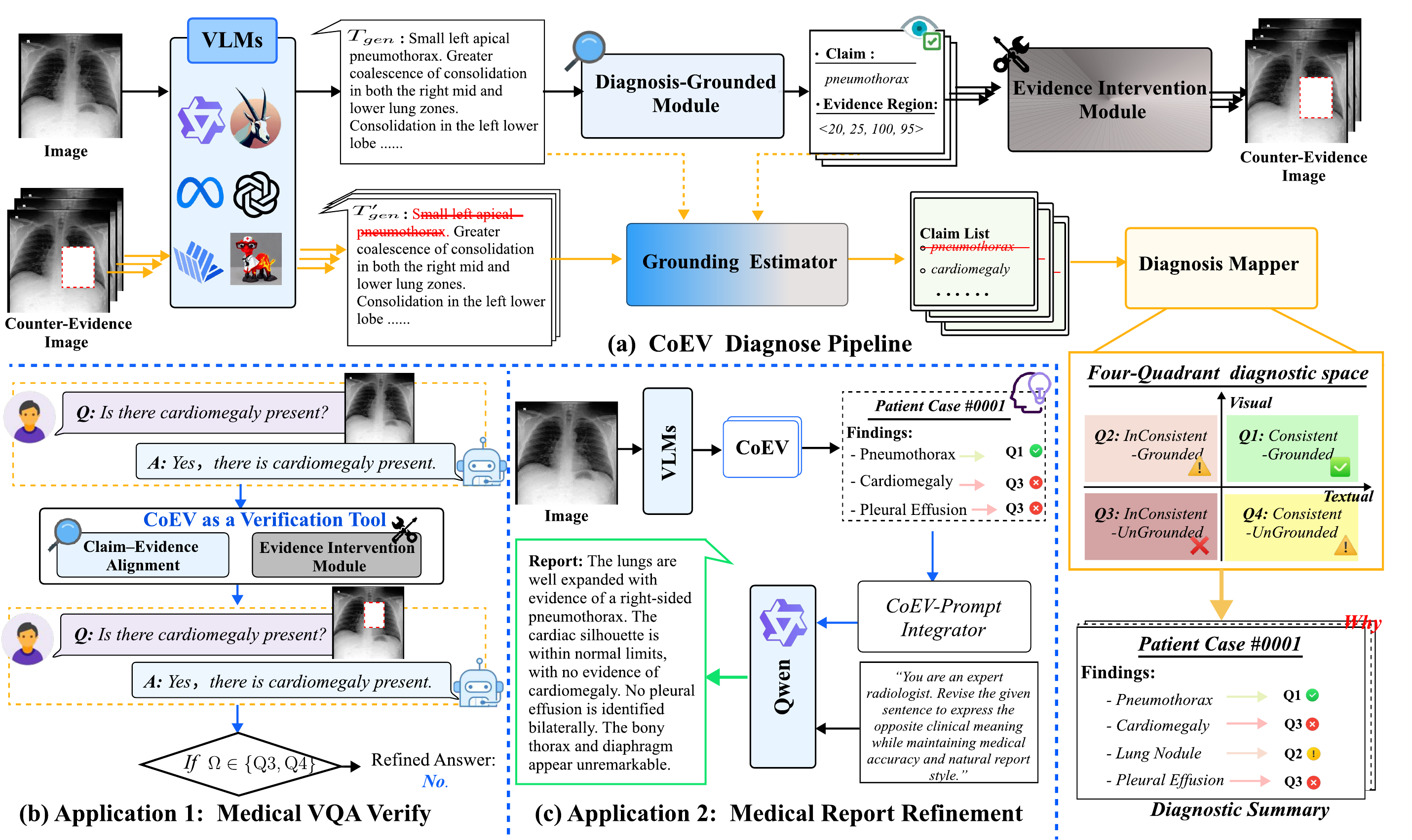} % 替换为你的图文件
    % \includegraphics[width=1\linewidth]{sec/Fig/F-2.png} % 替换为你的图文件

    % \caption{(a)\textbf{ Overview of the CoEV framework.} The {CoEV diagnostic process} performs counter-evidence verification and four-quadrant mapping to identify visually grounded and hallucinated claims from VLM outputs. (b) \textbf{CoEV for Med-VQA:} diagnostic signals are used to verify answer validity and identify hallucinated predictions. (c) \textbf{CoEV for Medical Report Refinement:} detected hallucinated sentences are refined through CoEV-guided rewriting.}
    \caption{(a) \textbf{CoEV Overview.} The {CoEV diagnostic process} performs counter-evidence verification and four-quadrant mapping to identify hallucinated claims. (b) Med-VQA Verification: Using diagnostic signals to validate answers and detect hallucinations. (c) Report Refinement: Rewriting hallucinated sentences using CoEV guidance.}
    \label{fig:F-2}
\end{figure}
\section{Method}
\subsection{Preliminaries}
\label{section-3-1}
\noindent \textbf{Hallucination Definition.} 
Building on~\cite{liu2024survey}, we define {hallucination} $y \in \{0, 1\}$ as a binary phenomenon conditioned on both textual and visual factors. Given a generated text $T_{\mathrm{gen}}$ and its corresponding image $I$, CoEV models the hallucination detection process as:
\begin{equation}
\label{halluciantion-define}
y = \mathbf{1}\big[(f(T_{\mathrm{gen}}) = 0) \lor (z(T_{\mathrm{gen}}, I) = 0)\big],
\end{equation}
where $f(\cdot)$ measures factual correctness and $z(\cdot)$ evaluates grounding consistency with visual evidence. Specifically, $y=1$ indicates a hallucination, which defined as a claim that is either factually incorrect ($f=0$) or insensitive to the masking of referenced visual regions ($z=0$). Conversely, a grounded claim ($y=0$) must be both factually accurate and visually dependent. Fig.~\ref{fig:F-2} illustrates how the CoEV integrate verification, reasoning, and visualization into an actionable diagnostic workflow for factual and visual correctness.

\noindent \textbf{Diagnosis-Grounded Module (DGM).} 
To operationalize the ``What \& Where'' linkage, the DGM establishes grounding between claims and visual evidence. Let $\mathcal{P}(\cdot)$ be a semantic parser that extracts positive factual claims $C_{\text{pos}}$ from generated text $T_{\mathrm{gen}}$:
\begin{equation}
C_{\text{pos}} = \{ c_i \mid \mathcal{P}(T_{\mathrm{gen}})[i] = 1 \},
\label{eq:DGM_extract}
\end{equation}
where label $1$ indicates a positive clinical finding. For each $c_i \in C_{\text{pos}}$, a radiology expert $\mathcal{G}$ localizes a visual region $B_i \subset I$ with confidence $s_i \in [0,1]$:
\begin{equation}
(B_i, s_i) = \mathcal{G}(I, c_i).
\label{eq:DGM_ground}
\end{equation}
We define a grounding indicator $\delta_i \in \{0, 1\}$ as the basis for subsequent probing: $\delta_i = 1$ if $s_i > 0$, and $\delta_i = 0$ otherwise. This structured output from DGM serves as the prerequisite for our proposed counterfactual verification.

\subsection{Counter-Evidence Consistency Evaluation (CECE).}
% \noindent \textbf{Counterfactual Probing and Attribution (CPA).}
\label{CPA}
The CECE module evaluates whether a generated claim is grounded in visual evidence or primarily driven by linguistic priors. CECE integrates two sub-processes: \textbf{{Evidence Intervention Module (EIM) and Grounding Estimator (GE)},} which jointly assess a claim’s generation depends on visual support or linguistic bias.
For each identified claim (\( \delta_i = 1 \)), a counter-evidence image is constructed by masking its corresponding visual region \( B_i \):
\begin{equation}
I'_i = \mathcal{M}(I, B_i),
\label{eq:I_i_prime}
\end{equation}

where $\mathcal{M}$ is an intervention operator that masks $B_i$ with black pixels while preserving the remaining image context.
The VLM is then re-evaluated on the modified input to obtain
\(
T'_i = g_{\theta}(I'_i).
\)

Let $\phi(c_i, T'_i) \in \{0, 1\}$ indicate if claim $c_i$ persists in the modified output $T'_i$. The grounding attribution $z_i \in \{0, 1\}$ is defined as:
\begin{equation}
z_i =
\begin{cases}
0, & \text{if } \phi(c_i, T'_i) = 1,\\
1, & \text{if } \phi(c_i, T'_i) = 0,\\
\end{cases}
\label{eq:z_i}
\end{equation}

Specifically, $z_i=1$ confirms visual grounding via evidence reliance, while $z_i=0$ reveals a prior-driven hallucination where the claim persists despite intervention. Unlike correctness-only methods, CECE explicitly verifies the \textit{causal} dependence of claims on their supporting visual context.% Unlike prior works that diagnose hallucination solely from textual correctness (\emph{``What''}), CPA explicitly tests the visual causality of each claim via counterfactual intervention.

% \subsection{Diagnosis Mapper (DM).}
% % \noindent \textbf{Diagnosis Mapper (DM).}
% \label{DM}
% The DM module integrates factual correctness and visual grounding to construct a unified diagnostic representation for each generated claim.  
% Given the factual correctness label \( f_i \in \{0,1\} \) obtained by comparing the generated claim with the ground-truth report \(T_{GT}\), and the grounding attribution variable \( z_i \in \{0,1\} \) derived from CECE, the mapping is defined as:
% \begin{equation}
% \mathcal{D}(f_i, z_i) =
% \begin{cases}
% Q_1, & f_i = 1, \ z_i = 1,\\
% Q_2, & f_i = 1, \ z_i = 0,\\
% Q_3, & f_i = 0, \ z_i = 1,\\
% Q_4, & f_i = 0, \ z_i = 0.
% \end{cases}
% \label{eq:D_fi_zi}
% \end{equation}

% Each claim $i$ is represented by its factual correctness $f_i$ and grounding consistency $z_i$, which jointly determine its position in the diagnostic space divided into four quadrants.  
% Following the hallucination definition in Sec~\ref{halluciantion-define}, we define the hallucination label $y_i$ such that $y_i = 0$ (grounded) if the diagnostic pair $(f_i, z_i) \in Q_1$, and $y_i = 1$ (hallucinated) if $(f_i, z_i) \in \{Q_2, Q_3, Q_4\}$, representing assertions that are either inconsistent with the image or lack sufficient visual evidence.
\subsection{Diagnosis Mapper (DM).}
\label{DM}
DM integrates factual correctness $f_i \in \{0, 1\}$ and grounding attribution $z_i \in \{0, 1\}$ into a unified diagnostic mapping $\mathcal{D}(f_i, z_i) \in \{Q_1, Q_2, Q_3, Q_4\}$:
\begin{equation}
    \mathcal{D}(f_i,z_i)=
        \begin{cases}
        \text{Q1},   & f_i=1, z_i=1 \\
        \text{Q2}, & f_i=0, z_i=1 \\
        \text{Q3}, & f_i=0, z_i=0 \\
        \text{Q4}, &f_i=1, z_i=0
        \end{cases}
\end{equation}
A claim is grounded ($y_i=0$) if $(f_i, z_i) \in Q_1$, and hallucinated ($y_i=1$) otherwise. These quadrants categorize assertions into {factual alignment} ($Q_1$), {language-driven bias} ($Q_2$), {visual misinterpretation} ($Q_3$), and {total failure} ($Q_4$). This taxonomy distinguishes medical inaccuracies from lack of visual dependence.
% Here, $y_i=0$ indicates a non-hallucinated claim, and $y_i=1$ indicates a hallucinated claim. This discrete formulation enables analyzable categorization of generated claims, revealing whether errors originate from factual inaccuracy, weak grounding, or language-driven bias.
\subsection{CoEV-Guided Correction Applications}
CoEV enables correction across two downstream tasks: (1) verification for close-ended Med-VQA and (2) refinement for open-ended medical reports.

\noindent\textbf{Med-VQA Verification.}
% For VQA-style tasks, the goal is to verify whether the model’s predicted answer is both factually correct and grounded in visual evidence. 
% Given an input image $I$ and a question $Q$, the baseline VLM produces an answer
% \(
% A = g_{\theta}(I, Q).
% \)
% This answer is then passed into CoEV:
% \begin{equation}
% \Omega = \mathcal{F}(I, Q, A)
% \label{eq:FIDES_output}
% \end{equation}
For a predicted answer $A = g_{\theta}(I, Q)$, CoEV evaluates its textual and visual integrity via $\Omega = \mathcal{F}(I, Q, A)$, ensuring the response is both accurate and visually grounded.
where $\mathcal{F}$ denote the CoEV,  and $\Omega \in \{Q_{1}, Q_{2}, Q_{3}, Q_{4}\}$.
We define a verification operator that produces a corrected or validated answer:
\begin{equation}
\hat{A} = A \oplus \mathbb{I}[\Omega \in {Q_{2}, Q_{3}}]
\label{eq:A_hat_impl}
\end{equation}

This mechanism enables {evidence-grounded verification} of close-ended predictions and refines visually unsupported or hallucinated answers.

\noindent\textbf{Report-level Refinement.} For MRG, CoEV enables sentence-level correction of hallucinated content. 
Let the baseline report be 
$R_{\text{gen}} = \{ s_1, s_2, \ldots, s_n \}$, 
and the diagnostic outcomes 
$\{(c_i, \Omega_i)\}$ be obtained by applying CoEV to each claim $c_i$ in the report.
Each sentence flagged as hallucinated ($\Omega_i \in \{Q_2, Q_3, Q_4\}$) is revised via a refinement operator $s'_i = \mathcal{R}(s_i, \texttt{p})$, which prompts the LLM to rewrite the statement based on validated visual evidence, ensuring the final report is both factually accurate and grounded.
% Each sentence flagged as hallucinated ($\Omega_i \in \{Q_{2}, Q_{3}\}$) 
% is revised via a refinement operator $\mathcal{R}$, which provides a \texttt{prompt} $\texttt{p}$ to the language model, instructing it to rewrite the statement based on visual evidence:
% \begin{equation}
% s'_i = \mathcal{R}(s_i, \texttt{p})
% \label{eq:s_prime}
% \end{equation}

The refined report is reconstructed by replacing hallucinated sentences while leaving others unchanged:
\begin{equation}
\hat{R} = 
\begin{cases}
s'_i, & \text{if } \Omega_i \in \{Q_{2}, Q_{3}\},\\
s_i, & \text{otherwise}.
\end{cases}
\label{eq:R_hat}
\end{equation}
This process leverages CoEV diagnostics to guide automatic correction, ensuring the final report aligns with factual and visual grounding consistency.
\section{Experiments}
\subsection{Experimental Setup}\noindent\textbf{Tasks and Datasets.} We evaluate CoEV on two medical vision-language tasks: \textbf{(1) Med-VQA}: Conducted on VQA-RAD~\cite{lau2018dataset},and MIMIC-VQA~\cite{bae2024mimic}, focusing on visually grounded clinical assertions.\textbf{(2) {MRG}}\textbf{: }Evaluated on MIMIC-CXR~\cite{johnson2019mimic} and IU-Xray~\cite{demner2015preparing} datasets using official test splits. We utilized the GREEN model~\cite{ostmeier2024green} to generate hallucination ground-truth labels by referencing gold-standard answers. 
\begin{table}[h]
\footnotesize
\centering
\caption{Hallucination detection performance (\%) across MRG and Med-VQA tasks. \textbf{PR}: PR-AUC; \textbf{ROC}: ROC-AUC. Best results are \textbf{bolded}, second-best are \underline{underline}.}
\label{tab:vqa-report-combined-percent}
% \setlength{\tabcolsep}{8.5pt}
% \setlength{\tabcolsep}{1pt}
% \resizebox{\columnwidth}{!}{%

\begin{tabular}{l cc cc cc cc}
\toprule
\multirow{2}{*}{\textbf{Method}} & \multicolumn{2}{c}{\textbf{MIMIC-CXR}} & \multicolumn{2}{c}{\textbf{IU-Xray}} & \multicolumn{2}{c}{\textbf{VQA-RAD}} & \multicolumn{2}{c}{\textbf{MIMIC-VQA}} \\
\cmidrule(lr){2-3} \cmidrule(lr){4-5} \cmidrule(lr){6-7} \cmidrule(lr){8-9}
 & PR $\uparrow$ & ROC $\uparrow$ & PR $\uparrow$ & ROC $\uparrow$ & PR $\uparrow$ & ROC $\uparrow$ & PR $\uparrow$ & ROC $\uparrow$ \\
\midrule
\multicolumn{9}{c}{\textbf{General-purpose Model: InternVL3-2B}~\cite{zhu2025internvl3}} \\
\midrule
RadFlag~\cite{zhang2024radflag} &70.61  &  54.02&77.02  &57.97 & \underline{65.32} & 54.09 & \underline{70.54} & 62.71 \\
SE~\cite{farquhar2024detecting} &76.96  & 52.54 & 78.62 &59.37 & 66.51 & 52.58 & 60.93 & 64.10 \\
VASE~\cite{liao2025vision} & \underline{81.66} & \underline{54.43} & \underline{80.00} & \underline{61.76} & 68.48 & \underline{58.11} & 62.79 & \underline{68.06} \\
\textbf{CoEV (Ours)} & \textbf{82.35} & \textbf{56.75} & \textbf{80.21} & \textbf{61.89}& \textbf{72.35} & \textbf{67.45} & \textbf{81.31} & \textbf{76.95} \\
\midrule
\multicolumn{9}{c}{\textbf{General-purpose Model: Qwen2-3B}~\cite{bai2023qwen}} \\
\midrule
RadFlag~\cite{zhang2024radflag}& 73.41 & 56.68 &87.56  &62.82 & {82.99} & \textbf{74.37} & \underline{79.91} & 61.08 \\
SE~\cite{farquhar2024detecting}  & \underline{78.74} & \underline{58.96}  & 89.20 &62.63 & 86.16 & 70.15 & 61.66 & 60.18 \\
VASE~\cite{liao2025vision}& 75.18 &  52.19&\textbf{89.92}  &\textbf{68.09} & \underline{89.71} & 61.81 & 71.01 & \underline{64.34} \\
\textbf{CoEV (Ours)} &\textbf{ 79.20 }  & \textbf{60.04 } & \underline{89.18} & \underline{64.61}& \textbf{92.16} & \underline{69.25} & \textbf{90.89} & \textbf{72.65} \\
\midrule
\multicolumn{9}{c}{\textbf{Medical-tuned Model: Lingshu-7B}~\cite{xu2025lingshu}} \\
\midrule
RadFlag~\cite{zhang2024radflag} & 83.29 &55.31  & 72.67 & 40.21& \underline{87.71} & 52.39 & \underline{84.10} & 61.37 \\
SE~\cite{farquhar2024detecting}& 84.73 & 59.48 & 77.53 &44.68 & 81.85 & 52.47 & 66.69 & 61.79 \\
VASE~\cite{liao2025vision} & \textbf{87.59}  & 61.87  & \underline{77.89} &\underline{49.66} & 87.45 & \underline{58.78} & 65.29 & \underline{65.51} \\
\textbf{CoEV (Ours)}& 87.10 & \underline{63.67} &\textbf{ 79.90} &\textbf{49.69 }& \textbf{89.73} & \textbf{65.63} & \textbf{85.10} & \textbf{70.60} \\
\midrule
\multicolumn{9}{c}{\textbf{Medical-tuned Model: LLaVA-Med}~\cite{li2023llava}} \\
\midrule
RadFlag~\cite{zhang2024radflag} & 91.53 & 53.13 & 70.61 & 54.02 & \underline{71.86} & 61.12 & \underline{79.54} & 54.11 \\
SE~\cite{farquhar2024detecting} & 92.76 & 54.94 & 66.70 & 52.54 & 63.00 & 64.34 & 62.40 & 62.00 \\
VASE~\cite{liao2025vision} & \textbf{94.26} & \underline{59.34} & \underline{75.94} & \underline{56.50} & 63.02 & \underline{67.01} & 62.57 & \underline{71.08} \\
\textbf{CoEV (Ours)} & \underline{93.94} & \textbf{61.97} & \textbf{76.17} & \textbf{59.95} & \textbf{86.57} & \textbf{68.94} & \textbf{72.06} & \textbf{76.12} \\
\bottomrule
\end{tabular}
% }
\end{table}

\noindent\textbf{Evaluation Metrics.} For hallucination detection, we employ PR-AUC and ROC-AUC. Report quality is assessed via diagnostic accuracy, following~\cite{bannur2024maira} (CheXpert-style $F_1$ scores: $\text{Mi-}F_{1\text{-}5}$ and $\text{Mi-}F_{1\text{-}14}$) and hallucination metrics: CHAIR-I/S~\cite{gu2025medvh} for instance/sentence-level errors, and MediHall~\cite{chen2024detecting} for clinically weighted penalties. VQA performance is measured by model accuracy. To assess generalizability, we benchmark CoEV across diverse VLMs, {spanning general-purpose and medical-tuned} architectures.

\begin{table}[h]
\footnotesize
\setlength{\tabcolsep}{2.8pt} 
\centering
\caption{Hallucination detection performance (\%). \textbf{F5/F14}: diagnostic $F_1$ scores; \textbf{C-I/C-S}: CHAIR-I/S $\downarrow$; \textbf{MH}: MediHall $\uparrow$. $\checkmark$ denotes our CoEV framework.}
\label{tab:hallucination-models}
% \resizebox{\columnwidth}{!}{
\setlength{\tabcolsep}{1pt}

\begin{tabular}{l c cc ccc cc ccc}
\toprule
\multirow{3}{*}{\textbf{Model}} & \multirow{3}{*}{\textbf{CoEV}} & \multicolumn{5}{c}{\textbf{MIMIC-CXR}} & \multicolumn{5}{c}{\textbf{IU-Xray}} \\
\cmidrule(lr){3-7} \cmidrule(lr){8-12}
& & \multicolumn{2}{c}{\textbf{Acc} $\uparrow$} & \multicolumn{3}{c}{\textbf{Hallu}} & \multicolumn{2}{c}{\textbf{Acc} $\uparrow$} & \multicolumn{3}{c}{\textbf{Hallu}} \\ 
\cmidrule(lr){3-4} \cmidrule(lr){5-7} \cmidrule(lr){8-9} \cmidrule(lr){10-12}
 & & $F_5$ & $F_{14}$ & C-I$\downarrow$ & C-S$\downarrow$ & MH$\uparrow$ & $F_5$ & $F_{14}$ & C-I$\downarrow$ & C-S$\downarrow$ & MH$\uparrow$ \\
\midrule
\multirow{2}{*}{{Qwen2-3B}} & \xmark & 16.95 & 16.46 & 69.68 & 16.25 & 42.80 & 12.50 & 17.41 & 90.80 & 18.82 & 70.73 \\
& \cellcolor{LightPurple}\cmark & \cellcolor{LightPurple}19.69 & \cellcolor{LightPurple}18.64 & \cellcolor{LightPurple}65.92 & \cellcolor{LightPurple}14.66 &\cellcolor{LightPurple} 43.80 & \cellcolor{LightPurple}32.78 & \cellcolor{LightPurple}57.63 & \cellcolor{LightPurple}86.93 & \cellcolor{LightPurple}14.32 & \cellcolor{LightPurple}72.74 \\

% \multirow{2}{*}{{Qwen2-7B}} & \xmark & 10.69 & 12.98 & 67.49 & 12.02 & 52.41 & 50.51 & 48.71 & 88.32 & 16.78 & 70.30 \\
%  & \cellcolor{LightPurple}\cmark & \cellcolor{LightPurple}10.99 & \cellcolor{LightPurple}13.67 & \cellcolor{LightPurple}62.03 & \cellcolor{LightPurple}9.71 & \cellcolor{LightPurple}55.34 & \cellcolor{LightPurple}51.44 & \cellcolor{LightPurple}57.40 & \cellcolor{LightPurple}85.55 & \cellcolor{LightPurple}12.95 & \cellcolor{LightPurple}72.54 \\

% \multirow{2}{*}{{InternVL-2B}} & \xmark & 15.97 & 22.57 & 58.85 & 13.35 & 65.72 & 12.90 & 25.28 & 74.68 & 7.93 & 82.91 \\
%  & \cellcolor{LightPurple}\cmark & \cellcolor{LightPurple}18.18 & \cellcolor{LightPurple}24.83 & \cellcolor{LightPurple}56.79 & \cellcolor{LightPurple}12.77 &\cellcolor{LightPurple} 68.73 &\cellcolor{LightPurple} 17.39 & \cellcolor{LightPurple}60.09 & \cellcolor{LightPurple}70.15 &\cellcolor{LightPurple} 7.80 &\cellcolor{LightPurple} 85.45 \\

\multirow{2}{*}{{InternVL-8B}} & \xmark & 49.80 & 40.30 & 49.10 & 21.33 & 70.78 & 24.84 & 56.64 & 61.33 & 7.71 & 86.24 \\
 &\cellcolor{LightPurple} \cmark & \cellcolor{LightPurple}64.96 & \cellcolor{LightPurple}47.89 & \cellcolor{LightPurple}38.67 & \cellcolor{LightPurple}14.63 & \cellcolor{LightPurple}72.24 & \cellcolor{LightPurple}30.65 & \cellcolor{LightPurple}57.86 & \cellcolor{LightPurple}49.47 & \cellcolor{LightPurple}6.09 & \cellcolor{LightPurple}87.71 \\

\multirow{2}{*}{{LLava-Med}} & \xmark & 18.76 & 21.12 & 74.06 & 27.08 & 34.01 & 22.83 & 43.41 & 86.98 & 23.77 & 76.50 \\
 & \cellcolor{LightPurple}\cmark &\cellcolor{LightPurple} 23.80 & \cellcolor{LightPurple}28.29 &\cellcolor{LightPurple} 72.92 & \cellcolor{LightPurple}23.43 &\cellcolor{LightPurple} 36.35 & \cellcolor{LightPurple}12.50 & \cellcolor{LightPurple}57.50 & \cellcolor{LightPurple}86.35 & \cellcolor{LightPurple}20.35 & \cellcolor{LightPurple}77.91 \\

% \multirow{2}{*}{{CXR-LLava}} & \xmark & 45.24 & 35.67 & 52.38 & 13.83 & 72.75 & 21.59 & 41.21 & 88.97 & 18.60 & 87.14 \\
%  &\cellcolor{LightPurple}\cmark &\cellcolor{LightPurple}65.99 &\cellcolor{LightPurple} 45.99 & \cellcolor{LightPurple}47.42 & \cellcolor{LightPurple}9.35 &\cellcolor{LightPurple} 79.79 &\cellcolor{LightPurple} 49.67 & \cellcolor{LightPurple}62.93 &\cellcolor{LightPurple} 82.51 &\cellcolor{LightPurple} 11.32 &\cellcolor{LightPurple} 89.90 \\

% \multirow{2}{*}{{XrayGPT}} & \xmark & 36.41 & 32.33 & 54.36& 13.42 & 71.06 & 15.11 & 43.31 & 90.09 & 18.99 & 71.75 \\
%  &\cellcolor{LightPurple} \cmark & \cellcolor{LightPurple}53.10 & \cellcolor{LightPurple}41.44 & \cellcolor{LightPurple}45.69 &\cellcolor{LightPurple} 11.22 &\cellcolor{LightPurple} 75.90 &\cellcolor{LightPurple} 35.29 &\cellcolor{LightPurple} 61.69 & \cellcolor{LightPurple}88.66 & \cellcolor{LightPurple}14.54 & \cellcolor{LightPurple}71.82 \\

% \multirow{2}{*}{{Maira-2}} & \xmark & 56.94 & 50.64 & 46.21 & 12.91 & 84.81 & 35.11 & 62.55 & 63.19 & 3.42 & 86.24 \\
%  &\cellcolor{LightPurple} \cmark &\cellcolor{LightPurple} 74.96 & \cellcolor{LightPurple}61.12 &\cellcolor{LightPurple} 34.58 &\cellcolor{LightPurple} 8.42 & \cellcolor{LightPurple}85.84 &\cellcolor{LightPurple} 48.53 & \cellcolor{LightPurple}66.19 &\cellcolor{LightPurple} 56.06 &\cellcolor{LightPurple} 2.87 & \cellcolor{LightPurple}86.52 \\

\multirow{2}{*}{{Lingshu-7B}} & \xmark & 30.35 & 24.93 & 47.35 & 9.53 & 85.38 & 25.97 & 57.97 & 56.45 & 1.90 & 90.88 \\
 & \cellcolor{LightPurple}\cmark &\cellcolor{LightPurple} 34.15 & \cellcolor{LightPurple}26.66 & \cellcolor{LightPurple}37.08 & \cellcolor{LightPurple}6.47 & \cellcolor{LightPurple}87.09 &\cellcolor{LightPurple} 30.77 & \cellcolor{LightPurple}57.55 & \cellcolor{LightPurple}41.30 & \cellcolor{LightPurple}10.79 & \cellcolor{LightPurple}91.70 \\
\bottomrule
\end{tabular}

\end{table}
\begin{table}[ht]
\centering
\footnotesize
\caption{Performance comparison on MIMIC-VQA and VQA-RAD. ``Ours'' indicates applying our refinement. Improvement ($\uparrow$) is shown in absolute \%.}
\label{label:Table-2}
\begin{tabular}{lcccccc}
\toprule
\multirow{2}{*}{\textbf{Model}} & \multicolumn{3}{c}{\textbf{MIMIC-VQA}} & \multicolumn{3}{c}{\textbf{VQA-RAD}} \\
\cmidrule(lr){2-4} \cmidrule(lr){5-7}
& w/o Ours & \cellcolor{LightPurple}w/ Ours & $\uparrow$ & w/o Ours & \cellcolor{LightPurple}w/ Ours & $\uparrow$ \\
\midrule
{Qwen2-3B} & 51.83 & \cellcolor{LightPurple}78.99 & \textbf{27.16} & 71.31 & \cellcolor{LightPurple}78.09 & \textbf{6.78} \\
% {Qwen2-7B} & 54.42 & \cellcolor{LightPurple}73.94 & \textbf{19.52} & 77.69 & \cellcolor{LightPurple}83.67 & \textbf{5.98} \\
% {InternVL-2B} & 56.25 & \cellcolor{LightPurple}69.04 & \textbf{12.79} & 73.31 & \cellcolor{LightPurple}78.88 & \textbf{5.57} \\
{InternVL-8B} & 60.38 & \cellcolor{LightPurple}74.13 & \textbf{13.75} & 76.89 & \cellcolor{LightPurple}82.47 & \textbf{5.58} \\
{LLaVA-Med}   & 68.56 & \cellcolor{LightPurple}78.99 & \textbf{10.43} & 71.35 & \cellcolor{LightPurple}85.66 & \textbf{14.31} \\
{Lingshu-7B}  & 64.42 & \cellcolor{LightPurple}78.22 & \textbf{13.80} & 78.09 & \cellcolor{LightPurple}91.24 & \textbf{13.15} \\
\bottomrule
\end{tabular}
\end{table}

\begin{table}[b]
\centering
\footnotesize
\caption{
Ablation of the contribution of CoEV’s two diagnostic axes on MIMIC-CXR.}  
% (1) \emph{textual consistency} ,  
% (2) \emph{visual consistency}.  }
% The full CoEV framework combines both axes to derive the four-quadrant diagnostic map.}
\label{tab:ablation_fides_modules_MAIRA}
\begin{tabular}{ c c c c c}
\toprule
  Textual Axis & Visual Axis & Precision  & Recall & F1-Score   \\
\midrule
   \textcolor{red}{\ding{55}} & \textcolor{red}{\ding{55}} & 0.5233 & 0.4095 & 0.5064   \\
   \textcolor{red}{\ding{55}} & \textcolor{green}{$\checkmark$} &0.6307 &0.4532 & 0.5460  \\
   \textcolor{green}{$\checkmark$} & \textcolor{red}{\ding{55}}&0.7474 &0.5042 & 0.5986 \\
   \textcolor{green}{$\checkmark$} & \textcolor{green}{$\checkmark$}&0.7750 &0.5046  & 0.6112  \\
\bottomrule
\end{tabular}
\end{table}

\begin{figure}[hbtp]
    \centering
    \includegraphics[width=0.85\linewidth]{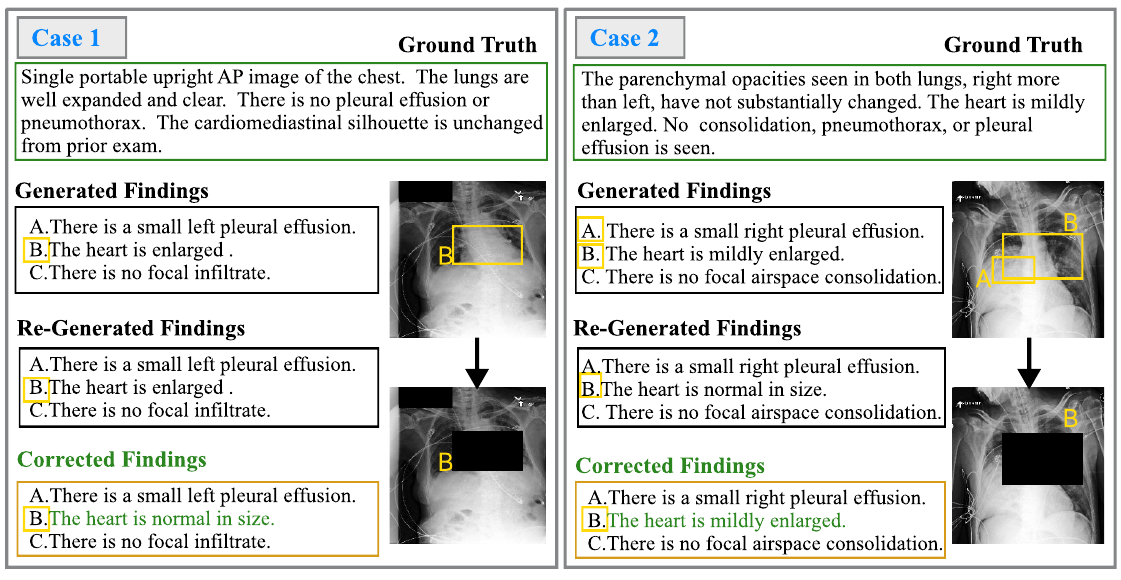} % 替换为你的图文件
    \caption{Representative cases. \textbf{Case 1 ($Q_2$)}: description persists after masking, indicating {ungrounded bias}. \textbf{Case 2 ($Q_1$)}: description disappears post-masking, reflecting {evidence-based reasoning}. CoEV refines reports by aligning claims with visual evidence.}
    % \caption{Two representative cases from \textbf{Lingshu} outputs under counter-evidence evaluation. The left case (1) shows a description that persists after counter-evidence masking, indicating an \textbf{ungrounded claim} corrected by CoEV. The right case (2) shows a description that disappears when the relevant region is masked, reflecting \textbf{evidence-grounded reasoning}. CoEV further refines the final output by aligning textual claims with verified visual evidence.}
    % Specialized medical models (e.g., Maira-2, XrayGPT) exhibit more visually grounded reasoning but still generate notable hallucinations, while general-purpose models rely heavily on non-causal reasoning.}
    \label{fig:F-5}
\end{figure}

\subsection{Results and Analyses} \noindent\textbf{Hallucination Detection Performance.} As shown in Table~\ref{tab:vqa-report-combined-percent}, CoEV consistently outperforms baselines across diverse model scales. Compared to the second-best method, CoEV achieves an average absolute improvement of {3.0\% in PR-AUC} and {3.1\% in ROC-AUC}. While SE exhibits instability in larger models and VASE yields moderate results, CoEV provides robust detection. Notably, on InternVL3-2B for MIMIC-VQA, CoEV improves PR-AUC and ROC-AUC by {10.77\%} and {8.89\%} respectively. This highlights that our framework effectively isolates ungrounded clinical claims where purely statistical metrics fail. {The performance gain across both open-set reporting and closed-set VQA tasks underscores CoEV's generalizability in safety-critical medical environments.}

\noindent\textbf{CoEV as a Post-hoc Refinement Tool.} Beyond detection, CoEV serves as an effective pipeline for downstream clinical tasks.
\textbf{(1) Medical Report Refinement:} As shown in Table~\ref{tab:hallucination-models}, CoEV significantly enhances diagnostic accuracy. On average, $F_{1\text{-}5}$ improves by {7.01\%} on MIMIC-CXR and {8.16\%} on IU-Xray. Notably, for InternVL-8B, $F_{1\text{-}5}$ increases from 49.80\% to 64.96\% ({+15.16\%}) on MIMIC-CXR. The reduction in hallucinations is evidenced by decreased CHAIR scores (e.g., C-I drops by {11.86\%} for IU-Xray on Qwen2-3B) and improved MediHall indices, confirming that CoEV successfully prunes unsupported clinical assertions while preserving factual findings.
\textbf{(2) Med-VQA Verification:} According to Table~\ref{label:Table-2}, CoEV yields a consistent mean accuracy gain of {13.04\%} across evaluated datasets. Smaller models exhibit the most pronounced gains, such as a {+27.16\%} absolute increase for Qwen2-3B on MIMIC-VQA. Even for specialized models like Lingshu-7B, CoEV provides steady improvements (avg. {+13.47\%}), underscoring its architecture-agnostic robustness. These results demonstrate that CoEV effectively mitigates hallucinations to improve the reliability of medical visual reasoning.

\subsection{Ablation Study}
% \noindent \textbf{Effect of the Two Diagnostic Axes.}
We ablate CoEV’s two core diagnostic dimensions:  
(1) the \emph{textual axis}, which evaluates textual consistency,  
and (2) the \emph{visual axis}, which evaluates whether a claim depends on the visual evidence.
% The full CoEV framework jointly incorporates both axes to form the four-quadrant diagnostic map.

As shown in Table~\ref{tab:ablation_fides_modules_MAIRA}, removing both axes yields the lowest performance with an F1-Score of 0.5064.  Using only the visual axis increases the F1-Score to 0.5460, indicating that visual dependence carries strong discriminative value for identifying ungrounded assertions.  Introducing only the textual axis improves precision by more than 20\%, yet it remains limited because visually unsupported but plausible statements cannot be distinguished.  
Combining both axes achieves the highest overall F1-Score of 0.6112. This joint model consistently improves precision and recall simultaneously, demonstrating that factual correctness and visual dependence provide complementary signals. This confirms that CoEV’s two-axis decomposition offers a meaningful and more discriminative structure for characterizing hallucination types.
Fig.~\ref{fig:F-5} illustrates CoEV's ability to distinguish and refine unsupported claims. In multi-disease scenarios, masking region A impacts only claim A, confirming that CoEV captures localized causal effects instead of global OOD noise, thus ensuring reliable refinement.

\section{Conclusion}
% We present \textbf{CoEV}, a unified framework for {Counter-Evidence Verification}, detecting and correcting hallucination in medical VLMs. Unlike prior methods that constrain outputs before inference, CoEV enables a plug-in-play diagnostic evaluation. It examines model assertions through visual evidence interventions, offering fine-grained explanations by assigning claims along textual and visual consistency in a four-quadrant diagnostic map. This design enables accurate hallucination detection across diverse VLMs. Extensive experiments on MRG and Med-VQA benchmarks demonstrate that CoEV effectively detects hallucinations and improves factual accuracy and reduces hallucinations across diverse VLMs. Overall, CoEV provides a framework to detect and correct hallucinations in VLMs. Looking forward, we aim to extend CoEV to correct  visual inconsistencies in addition to textual hallucinations, further improving the reliability of medical VLMs.
We present \textbf{CoEV}, a unified \textbf{Counter-Evidence Verification} framework for detecting and correcting hallucinations in medical VLMs. Unlike prior methods that constrain outputs during generation, CoEV enables a plug-and-play post-hoc evaluation that integrates seamlessly into existing clinical workflows. By intervening on visual evidence and mapping claims onto a four-quadrant diagnostic space, CoEV provides fine-grained, interpretable explanations that help clinicians trace the origin of model errors. Extensive experiments on MRG and Med-VQA benchmarks demonstrate that CoEV consistently outperforms baselines, significantly enhancing factual accuracy while reducing hallucination rates across various architectures. Ultimately, CoEV establishes a robust foundation for evidence-based reliability in safety-critical AI. Looking forward, we aim to extend this framework to address multi-modal inconsistencies and explore its deployment in real-time clinical decision support systems.
\bibliographystyle{splncs04}
\bibliography{main}

@String(AAAI = {AAAI})

@inproceedings{ostmeier2024green,
  title={Green: Generative radiology report evaluation and error notation},
  author={Ostmeier, Sophie and Xu, Justin and Chen, Zhihong and Varma, Maya and Blankemeier, Louis and Bluethgen, Christian and Md, Arne Edward Michalson and Moseley, Michael and Langlotz, Curtis and Chaudhari, Akshay S and others},
  booktitle={Findings of the association for computational linguistics: EMNLP 2024},
  pages={374--390},
  year={2024}
}

@article{liu2024survey,
  title={A survey on hallucination in large vision-language models},
  author={Liu, Hanchao and Xue, Wenyuan and Chen, Yifei and Chen, Dapeng and Zhao, Xiutian and Wang, Ke and Hou, Liping and Li, Rongjun and Peng, Wei},
  journal={arXiv preprint arXiv:2402.00253},
  year={2024}
}

@article{zou2025uncertainty,
  title={Uncertainty-aware medical diagnostic phrase identification and grounding},
  author={Zou, Ke and Bai, Yang and Liu, Bo and Chen, Yidi and Chen, Zhihao and Zhou, Yang and Yuan, Xuedong and Wang, Meng and Shen, Xiaojing and Cao, Xiaochun and others},
  journal={IEEE Transactions on Pattern Analysis and Machine Intelligence},
  year={2025},
  publisher={IEEE}
}

@inproceedings{liu2025gemex,
  title={Gemex: A large-scale, groundable, and explainable medical vqa benchmark for chest x-ray diagnosis},
  author={Liu, Bo and Zou, Ke and Zhan, Li-Ming and Lu, Zexin and Dong, Xiaoyu and Chen, Yidi and Xie, Chengqiang and Cao, Jiannong and Wu, Xiao-Ming and Fu, Huazhu},
  booktitle={Proceedings of the IEEE/CVF International Conference on Computer Vision},
  pages={21310--21320},
  year={2025}
}

@article{zou2024medrg,
  title={Medrg: Medical report grounding with multi-modal large language model},
  author={Zou, Ke and Bai, Yang and Chen, Zhihao and Zhou, Yang and Chen, Yidi and Ren, Kai and Wang, Meng and Yuan, Xuedong and Shen, Xiaojing and Fu, Huazhu},
  journal={arXiv e-prints},
  pages={arXiv--2404},
  year={2024}
}

@article{zhang2023biomedclip,
  title={Biomedclip: a multimodal biomedical foundation model pretrained from fifteen million scientific image-text pairs},
  author={Zhang, Sheng and Xu, Yanbo and Usuyama, Naoto and Xu, Hanwen and Bagga, Jaspreet and Tinn, Robert and Preston, Sam and Rao, Rajesh and Wei, Mu and Valluri, Naveen and others},
  journal={arXiv:2303.00915},
  year={2023}
}

@article{farquhar2024detecting,
  title={Detecting hallucinations in large language models using semantic entropy},
  author={Farquhar, Sebastian and Kossen, Jannik and Kuhn, Lorenz and Gal, Yarin},
  journal={Nature},
  volume={630},
  number={8017},
  pages={625--630},
  year={2024},
  publisher={Nature Publishing Group UK London}
}

@article{sellergren2025medgemma,
  title={Medgemma technical report},
  author={Sellergren, Andrew and Kazemzadeh, Sahar and Jaroensri, Tiam and Kiraly, Atilla and Traverse, Madeleine and Kohlberger, Timo and Xu, Shawn and Jamil, Fayaz and Hughes, C{\'\i}an and Lau, Charles and others},
  journal={arXiv preprint arXiv:2507.05201},
  year={2025}
}

@article{thawkar2023xraygpt,
  title={Xraygpt: Chest radiographs summarization using medical vision-language models},
  author={Thawkar, Omkar and Shaker, Abdelrahman and Mullappilly, Sahal Shaji and Cholakkal, Hisham and Anwer, Rao Muhammad and Khan, Salman and Laaksonen, Jorma and Khan, Fahad Shahbaz},
  journal={arXiv preprint arXiv:2306.07971},
  year={2023}
}

@article{li2023llava,
  title={Llava-med: Training a large language-and-vision assistant for biomedicine in one day},
  author={Li, Chunyuan and Wong, Cliff and Zhang, Sheng and Usuyama, Naoto and Liu, Haotian and Yang, Jianwei and Naumann, Tristan and Poon, Hoifung and Gao, Jianfeng},
  journal={Advances in Neural Information Processing Systems},
  volume={36},
  pages={28541--28564},
  year={2023}
}

@article{zhu2025internvl3,
  title={Internvl3: Exploring advanced training and test-time recipes for open-source multimodal models},
  author={Zhu, Jinguo and Wang, Weiyun and Chen, Zhe and Liu, Zhaoyang and Ye, Shenglong and Gu, Lixin and Tian, Hao and Duan, Yuchen and Su, Weijie and Shao, Jie and others},
  journal={arXiv preprint arXiv:2504.10479},
  year={2025}
}

@article{bai2023qwen,
  title={Qwen-vl: A frontier large vision-language model with versatile abilities},
  author={Bai, Jinze and Bai, Shuai and Yang, Shusheng and Wang, Shijie and Tan, Sinan and Wang, Peng and Lin, Junyang and Zhou, Chang and Zhou, Jingren},
  journal={arXiv preprint arXiv:2308.12966},
  volume={1},
  number={2},
  pages={3},
  year={2023}
}

@inproceedings{liao2025vision,
  title={Vision-amplified semantic entropy for hallucination detection in medical visual question answering},
  author={Liao, Zehui and Hu, Shishuai and Zou, Ke and Fu, Huazhu and Zhen, Liangli and Xia, Yong},
  booktitle={International Conference on Medical Image Computing and Computer-Assisted Intervention},
  pages={669--679},
  year={2025},
  organization={Springer}
}

@article{moslonka2025learned,
  title={Learned Hallucination Detection in Black-Box LLMs using Token-level Entropy Production Rate},
  author={Moslonka, Charles and Randrianarivo, Hicham and Garnier, Arthur and Malherbe, Emmanuel},
  journal={arXiv preprint arXiv:2509.04492},
  year={2025}
}

@article{zhang2024radflag,
  title={Radflag: A black-box hallucination detection method for medical vision language models},
  author={Zhang, Serena and Sambara, Sraavya and Banerjee, Oishi and Acosta, Julian and Fahrner, L John and Rajpurkar, Pranav},
  journal={arXiv preprint arXiv:2411.00299},
  year={2024}
}

@inproceedings{gunjal2024detecting,
  title={Detecting and preventing hallucinations in large vision language models},
  author={Gunjal, Anisha and Yin, Jihan and Bas, Erhan},
  booktitle={Proceedings of the AAAI Conference on Artificial Intelligence},
  volume={38},
  number={16},
  pages={18135--18143},
  year={2024}
}

@article{xu2025lingshu,
  title={Lingshu: A Generalist Foundation Model for Unified Multimodal Medical Understanding and Reasoning},
  author={Xu, Weiwen and Chan, Hou Pong and Li, Long and Aljunied, Mahani and Yuan, Ruifeng and Wang, Jianyu and Xiao, Chenghao and Chen, Guizhen and Liu, Chaoqun and Li, Zhaodonghui and others},
  journal={arXiv preprint arXiv:2506.07044},
  year={2025}
}

@article{an2022attention,
  title={Attention map-guided visual explanations for deep neural networks},
  author={An, Junkang and Joe, Inwhee},
  journal={Applied Sciences},
  volume={12},
  number={8},
  pages={3846},
  year={2022},
  publisher={MDPI}
}

@article{gu2025medvh,
  title={Medvh: Toward systematic evaluation of hallucination for large vision language models in the medical context},
  author={Gu, Zishan and Chen, Jiayuan and Liu, Fenglin and Yin, Changchang and Zhang, Ping},
  journal={Advanced Intelligent Systems},
  pages={2500255},
  year={2025},
  publisher={Wiley Online Library}
}

@inproceedings{xiao2025detecting,
  title={Detecting and mitigating hallucination in large vision language models via fine-grained ai feedback},
  author={Xiao, Wenyi and Huang, Ziwei and Gan, Leilei and He, Wanggui and Li, Haoyuan and Yu, Zhelun and Shu, Fangxun and Jiang, Hao and Zhu, Linchao},
  booktitle={Proceedings of the AAAI Conference on Artificial Intelligence},
  volume={39},
  number={24},
  pages={25543--25551},
  year={2025}
}

@article{chen2024detecting,
  title={Detecting and evaluating medical hallucinations in large vision language models},
  author={Chen, Jiawei and Yang, Dingkang and Wu, Tong and Jiang, Yue and Hou, Xiaolu and Li, Mingcheng and Wang, Shunli and Xiao, Dongling and Li, Ke and Zhang, Lihua},
  journal={arXiv preprint arXiv:2406.10185},
  year={2024}
}

@article{bannur2024maira,
  title={Maira-2: Grounded radiology report generation},
  author={Bannur, Shruthi and Bouzid, Kenza and Castro, Daniel C and Schwaighofer, Anton and Thieme, Anja and Bond-Taylor, Sam and Ilse, Maximilian and P{\'e}rez-Garc{\'\i}a, Fernando and Salvatelli, Valentina and Sharma, Harshita and others},
  journal={arXiv preprint arXiv:2406.04449},
  year={2024}
}

@article{johnson2019mimic,
  title={MIMIC-CXR, a de-identified publicly available database of chest radiographs with free-text reports},
  author={Johnson, Alistair EW and Pollard, Tom J and Berkowitz, Seth J and Greenbaum, Nathaniel R and Lungren, Matthew P and Deng, Chih-ying and Mark, Roger G and Horng, Steven},
  journal={Scientific data},
  volume={6},
  number={1},
  pages={317},
  year={2019},
  publisher={Nature Publishing Group UK London}
}

@misc{bae2024mimic,
  title={Mimic-ext-mimic-cxr-vqa: A complex, diverse, and large-scale visual question answering dataset for chest x-ray images},
  author={Bae, Seongsu and Kyung, Daeun and Ryu, Jaehee and Cho, Eunbyeol and Lee, Gyubok and Kweon, Sunjun and Oh, Jungwoo and Ji, Lei and Chang, Eric and Kim, Tackeun and others},
  year={2024},
  publisher={PhysioNet}
}

@article{lau2018dataset,
  title={A dataset of clinically generated visual questions and answers about radiology images},
  author={Lau, Jason J and Gayen, Soumya and Ben Abacha, Asma and Demner-Fushman, Dina},
  journal={Scientific data},
  volume={5},
  number={1},
  pages={1--10},
  year={2018},
  publisher={Nature Publishing Group}
}

@article{demner2015preparing,
  title={Preparing a collection of radiology examinations for distribution and retrieval},
  author={Demner-Fushman, Dina and Kohli, Marc D and Rosenman, Marc B and Shooshan, Sonya E and Rodriguez, Laritza and Antani, Sameer and Thoma, George R and McDonald, Clement J},
  journal={Journal of the American Medical Informatics Association},
  volume={23},
  number={2},
  pages={304--310},
  year={2015},
  publisher={Oxford Academic}
}

@article{raghavan2024attention,
  title={Attention guided grad-CAM: an improved explainable artificial intelligence model for infrared breast cancer detection},
  author={Raghavan, Kaushik and B, Sivaselvan and v, Kamakoti},
  journal={Multimedia Tools and Applications},
  volume={83},
  number={19},
  pages={57551--57578},
  year={2024},
  publisher={Springer}
}

@inproceedings{papineni2002bleu,
  title={Bleu: a method for automatic evaluation of machine translation},
  author={Papineni, Kishore and Roukos, Salim and Ward, Todd and Zhu, Wei-Jing},
  booktitle={Proceedings of the 40th annual meeting of the Association for Computational Linguistics},
  pages={311--318},
  year={2002}
}

@inproceedings{lin2004rouge,
  title={Rouge: A package for automatic evaluation of summaries},
  author={Lin, Chin-Yew},
  booktitle={Text summarization branches out},
  pages={74--81},
  year={2004}
}

\end{document}